\documentclass{ecai} 

\usepackage{latexsym}
\usepackage{amssymb}
\usepackage{amsmath}
\usepackage{amsthm}
\usepackage{booktabs}
\usepackage{enumitem}
\usepackage{graphicx}
\usepackage{color}

\usepackage[hidelinks]{hyperref}
\usepackage{cleveref}
\usepackage{amsmath,amssymb,amsfonts}
\usepackage{algorithmic}
\usepackage{graphicx}
\usepackage{textcomp}
\usepackage{xcolor}
\usepackage{subfig}
\usepackage{float}
\usepackage{amsmath}
\usepackage{amsthm}
\usepackage{booktabs}  %
\usepackage{algorithmic}
\usepackage{multirow}

\usepackage{csquotes} 
\MakeOuterQuote{"}

\newcommand{\BibTeX}{B\kern-.05em{\sc i\kern-.025em b}\kern-.08em\TeX}

\begin{document}


\begin{frontmatter}


\paperid{123} 


\title{Tabular Data Adapters: Improving Outlier Detection for Unlabeled Private Data}


\author[A,B]{\fnms{Dayananda}~\snm{Herurkar}\thanks{Corresponding Author. Email: dayananda.herurkar@dfki.de}}
\author[A,C]{\fnms{Jörn}~\snm{Hees}}
\author[D]{\fnms{Vesselin}~\snm{Tzvetkov}}
\author[A,B]{\fnms{Andreas}~\snm{Dengel}}

\address[A]{German Research Center for Artificial Intelligence (DFKI), Kaiserslautern, Germany}
\address[B]{RPTU Kaiserslautern-Landau, Germany}
\address[C]{Hochschule Bonn-Rhein-Sieg (H-BRS), Germany}
\address[D]{Google}


\begin{abstract}
The remarkable success of Deep Learning approaches is often based and demonstrated on large public datasets.
However, when applying such approaches to internal, private datasets, one frequently faces challenges arising from structural differences in the datasets, domain shift, and the lack of labels.
In this work, we introduce Tabular Data Adapters (TDA), a novel method for generating soft labels for unlabeled tabular data in outlier detection tasks. 
By identifying statistically similar public datasets and transforming private data (based on a shared autoencoder) into a format compatible with state-of-the-art public models, our approach enables the generation of weak labels.
It thereby can help to mitigate the cold start problem of labeling by basing on existing outlier detection models for public datasets.
In experiments on 50 tabular datasets across different domains, we demonstrate that our method is able to provide more accurate annotations than baseline approaches while reducing computational time. 
Our approach offers a scalable, efficient, and cost-effective solution, to bridge the gap between public research models and real-world industrial applications.
\end{abstract}

\end{frontmatter}


\section{Introduction}

Tabular data plays a vital role in numerous industries, supporting critical applications in sectors such as healthcare, manufacturing, and retail. 
In recent years, machine learning (ML) and deep learning (DL) techniques have emerged as powerful tools for solving complex problems in these domains \cite{OZBAYOGLU2020106384, AI-in-Finance}. 
One prominent example is fraud detection, which has gained significant attention due to the rise in fraudulent activities \cite{Fin-Fraud1, Fin-Fraud2}. 
Despite the potential of ML and DL, a common challenge faced by industries is the nature of the data they handle. 
More often than not, industrial datasets are structured differently than public research datasets, making it difficult to apply advanced techniques effectively \cite{fang2023}. 
For data owners aiming to build ML or outlier detection (OD) applications, the lack of labeled data poses significant challenges. 
These challenges include cold start problems, limited guidance on which ML approach might be useful, and the need for extensive trial-and-error approaches, such as hyperparameter tuning and manual labeling, both of which are time-consuming and costly.

Although state-of-the-art outlier detection models have been developed based on extensive research and public datasets, these models are often not directly applicable to private, unlabeled datasets. 
A common but flawed approach is to identify a publicly accessible model trained on a task similar to that of the unlabeled dataset and then attempt to apply it directly. 
Unfortunately, this method tends to be ineffective due to discrepancies in dataset characteristics, such as feature distributions, dimensionality, and data structure. 
Moreover, it is often impractical to find an exact match between public datasets and the proprietary datasets in use across different industries. 
As a result, despite the availability of high-performing public models, their utility in solving real-world problems is limited, particularly in scenarios where labeled data is scarce or nonexistent.

\begin{figure*}
\centering
\includegraphics[width=\linewidth]{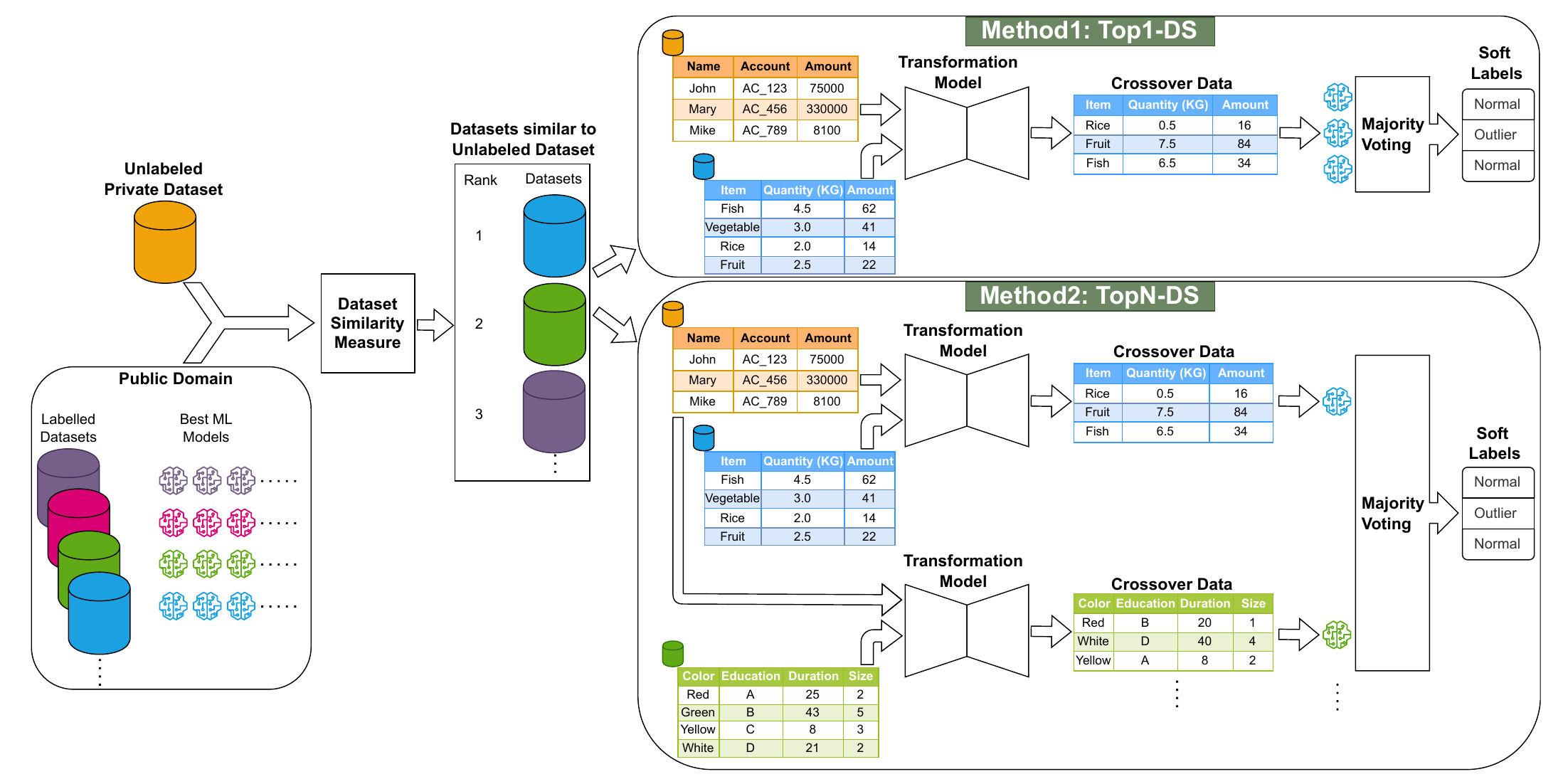}
\caption{Overview of our proposed soft labeling approach.
The private dataset is compared by a "dataset similarity measure" with an index of existing public domain datasets (for each of them the best working ML/OD models are known), resulting in the top $n$ most similar datasets.
\textbf{Method1: Top1-DS} performs a "Dataset Transformation" of the private dataset to the top $1$ most similar public dataset to align their formats. Predictions are generated using the known best $m$ optimized ML models for the top $1$ selected public dataset, and soft labels are created through majority voting.
\textbf{Method 2: TopN-DS} transforms the private dataset into the formats of the top $n$ similar public datasets. The known best-performing model from each transformed dataset is used to generate predictions, which are fused via majority voting.
Both methods aim to efficiently and accurately generate soft labels for unlabeled private datasets. \vspace{2mm}}
\label{fig:m1_and_m2}
\end{figure*}

To address this critical gap, we propose a novel approach (see \autoref{fig:m1_and_m2}) for generating soft labels for unlabeled tabular data in outlier detection. 
Rather than directly transferring models from public datasets to private datasets, our method first focuses on identifying the most statistically similar publicly available dataset.
The underlying hypothesis is that if a model performs well on a public dataset with statistical characteristics similar to those of the private dataset, then it is likely to perform reasonably well on the private data as well, even if the latter is unlabeled.
Then, by using the "Dataset Transformation" \cite{Herurkar-ijcnn23} technique, we facilitate the adaptation of public models to private datasets.
This technique uses self-supervised transcoders based on autoencoder pairs that share a latent (bottleneck) representation, allowing for the transformation of private data into a form compatible with the public model’s input space. 
By doing so, we enable the application of state-of-the-art models trained on public datasets to generate soft labels for private data. 
These soft labels can then be mapped back to the original unlabeled dataset, providing a starting point for training ML models.

Our approach is applicable to a wide range of domains and can deal with differing data structures, sample counts and feature dimensions. 
Through comprehensive experiments conducted across 50 datasets from varying fields, we demonstrate that our approach generates more accurate soft labels compared to baseline methods. 
Moreover, our technique requires less compute time to produce these labels, than baselines relying on individual hyperparameter tuning, making it highly practical for real-world applications where time and resources are limited.
In summary, our proposed method offers a promising solution to one of the most pressing challenges in industrial applications: the effective utilization of unlabeled data to mitigate the cold start problem of ML.
By leveraging publicly available datasets and models in a novel way, we create a bridge between academic research and practical industry needs, facilitating the application of ML and DL techniques in settings where labeled data is scarce or unavailable.
Our key contributions are as follows:
\begin{itemize}
    \item We introduce a novel, dimension-independent Dataset Similarity Measure to identify public datasets that are statistically similar to a given private, unlabeled dataset.
    \item We propose a complete, end-to-end soft labeling pipeline tailored for tabular outlier detection, enabling practical application in cold-start, unlabeled scenarios.
    \item We develop two distinct soft labeling strategies: Top1-DS, which leverages the most similar public dataset and its best-performing models, and TopN-DS, which leverages multiple similar datasets and fuses predictions from their top model for improved robustness and accuracy.
\end{itemize}

The remainder of this work is structured as follows: Section 2 provides a review of relevant literature, highlighting gaps in the domain. 
In Section 3, we detail our approach to soft labeling in the tabular dataset. 
The experimental setup, OD models, datasets used, and evaluation measures are described in Section 4, while Section 5 presents the results and comparisons. 
We conclude in Section 6, summarizing our main findings and identifying opportunities for future research.

\section{Related Work}
\subsection{Unsupervised OD Approaches for Tabular Data:}

Numerous techniques have been developed to identify irregular patterns in tabular data for outlier detection. 
Goldstein et al. \cite{goldstein} conducted a comprehensive analysis of 19 unsupervised outlier detection algorithms, spanning clustering, distance-based, density-based, and statistical methods, applied to 10 standard datasets. 
Their work highlights the strengths and limitations of these approaches in handling multivariate data. 
Similarly, Zhao et al. \cite{han2022adbench} analyzed 30 OD algorithms on 57 datasets and developed PyOD, an easy-to-use Python library for detecting anomalies in multivariate data. 
Outlier detection has long been a focal point of research, particularly in sectors like finance \cite{Fin-Fraud3}. 
The growing availability of tabular data in financial applications has spurred advancements in OD methods. 
The rise of deep learning (DL) has introduced novel approaches for analyzing tabular data to tackle complex challenges \cite{Borisov-2022}. 
Such as anomaly detection in accounting datasets \cite{RECol, Marco2}, interpretation of outliers in financial data \cite{Timur}, transforming outliers across different tabular formats \cite{Herurkar-ijcnn23}, and distillation of tabular data for OD \cite{Tab-Distillation}.
Emerging techniques such as federated outlier detection \cite{Fin-Fed-OD}, fraud pattern modeling \cite{wedge2017solving}, and enhancements to anti-money laundering systems \cite{Zahra, Ebberth} have demonstrated noteworthy progress, particularly in combating credit card fraud \cite{Pumsirirat2018, Zahra}. 
Reconstruction-based methods are another prominent category, operating on the premise that outliers are harder to reconstruct from lower-dimensional projections \cite{chalapathy2019deep, OZBAYOGLU2020106384}. 
For instance, DAGMM \cite{DAGMM} integrates density estimation, dimensionality reduction, and reconstruction error to identify anomalies in low-density regions.
Traditional techniques like Gaussian Mixture Models (GMMs) and K-means clustering \cite{k-means} remain popular, but often struggle with high-dimensional data due to the curse of dimensionality. 
On the other hand, one-class classification methods, like one-class SVM \cite{deep-one-class}, define a decision boundary to separate normal data from outliers, offering a complementary perspective for OD.

\subsection{Cross Domain Transformation and Data Sharing:}

Access to private datasets in sensitive fields like finance and healthcare has become increasingly restricted due to data privacy regulations such as GDPR. 
Common solutions include anonymization \cite{deanony1}, synthetic data generation \cite{Pei2006ASD, vaeout}, and differential privacy (DP) \cite{dp}. 
Anonymization removes personal identifiers to maintain privacy but often leads to information loss and remains vulnerable to de-anonymization attacks \cite{deanony2}. 
Synthetic data generation, using methods like variational autoencoders (VAE) and generative adversarial networks (GANs), has shown promise in creating realistic datasets, although most advancements have focused on image data rather than tabular data. 
DP, which introduces noise to protect individual records, is effective but can degrade data quality and model accuracy, particularly in small or customized datasets. 
A novel technique called Dataset Transformation \cite{Herurkar-ijcnn23} has recently emerged, enabling the transformation of private datasets into the format of public datasets. 
This method facilitates adaptation between datasets with different dimensions and feature types, making it practical in scenarios where proprietary and public datasets differ significantly.


Our study differs from prior approaches in several key aspects. 
By integrating a novel Dataset Similarity Measure with Dataset Transformation techniques, we directly tackle the challenge of label scarcity in private datasets. 
To the best of our knowledge, this is the first approach that enables practical and time-efficient soft labeling for proprietary tabular data by systematically leveraging a repository of pre-existing, optimized public datasets and models.

\section{Approach}
In this section, we outline the main components of our soft labeling method and its two variations.
An overview of our approach can be found in \autoref{fig:m1_and_m2}.

\subsection{Components of our Soft Labeling Model:}
\begin{itemize}
    \item \textbf{Dataset Similarity Measure:} 
    This component computes a heuristic for the similarity between two tabular datasets by comparing their information and redundancy using Principal Component Analysis (PCA).
    PCA is widely used for dimensionality reduction while preserving the most significant information from the original dataset \cite{abdi2010principal}.
    In our method, we apply PCA to the private dataset and calculate the mean reconstruction error by comparing the original dataset with its PCA-reconstructed version.
    This mean reconstruction error quantifies the amount of information lost during compression, reflecting the dataset’s complexity.
    The process is repeated for varying numbers of principal components (PCs), from 1 to 100, and the resulting mean reconstruction errors are plotted (\autoref{fig:ds_sim_pca}).
    The same procedure is applied to a public dataset, and the results can be plotted into the same graph.
    To heuristically quantify the difference between the two datasets, we calculate the sum of absolute differences (SAD) between the reconstruction errors of the private and public dataset. 
    To find the most similar public datasets, we rank them descending by their 
    SAD to the private dataset.
    The advantage of this measure over existing similarity methods is that it operates independently of feature dimensionality and does not rely on label information, making it broadly applicable across heterogeneous tabular datasets.
\begin{figure}
\centering
\includegraphics[width=\linewidth]{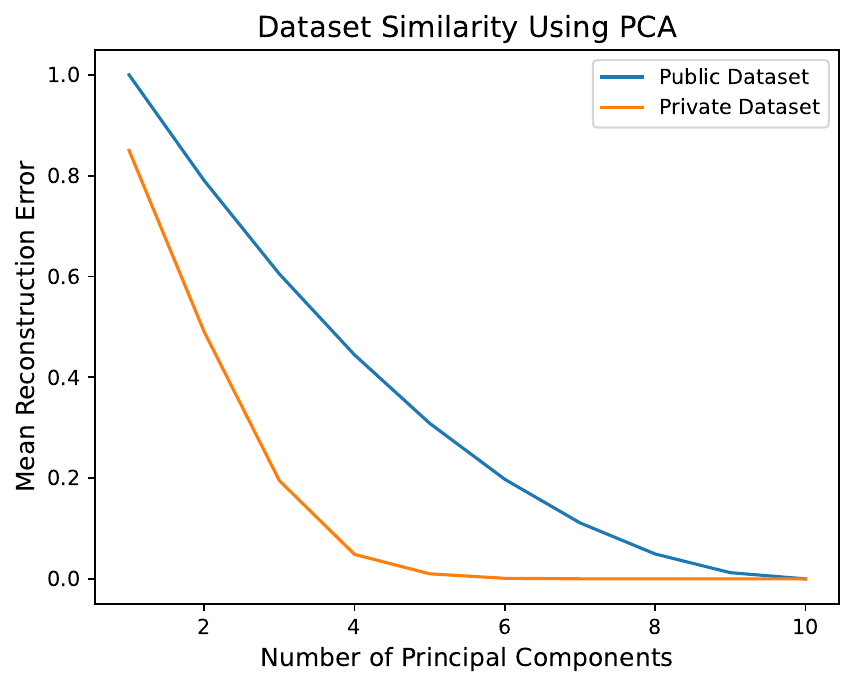}
\caption{Visualization of the mean reconstruction errors for private and public datasets using PCA. The x-axis represents the number of principal components (PCs), and the y-axis shows the corresponding mean reconstruction errors. The plotted curves highlight the complexity of compressing information of each dataset. The difference between the curves, quantified by the sum of absolute differences (SAD) of reconstruction errors, is used as the 
metric to measure dataset similarity.}
\label{fig:ds_sim_pca}
\end{figure}

    \item \textbf{Dataset Transformation:} 
    This component (\autoref{fig:ds_transformation_model}) leverages an autoencoder-based neural network to transform samples from private datasets into public datasets for outlier detection, following \cite{Herurkar-ijcnn23}. 
    The model consists of two types of layers: 
    \begin{itemize}
        \item Dedicated Layers: These include separate encoder-decoder pairs for the private (Encoder$_{prv}$-Decoder$_{prv}$) and public datasets (Encoder$_{pub}$-Decoder$_{pub}$), responsible for learning dataset-specific patterns.
        \item Shared Layers ($m_{\theta}$): Common to both datasets, these layers capture patterns applicable across both private and public data.
    \end{itemize}
    During the training phase, batches from either the private ($X_{prv}$) or public ($X_{pub}$) dataset are alternately fed into the model where $X_{prv}$ and $X_{pub}$ comprises a set of attributes $d\in{\{1,...,D_{prv}\}}$ and $d\in{\{1,...,D_{pub}\}}$, respectively. The respective encoder compresses the data into lower dimensions, which is then processed by the shared layers. The decoder reconstructs the original data from the shared layer outputs. This process allows the model to learn both specific and common patterns from both datasets.
    \begin{equation}
	\hat{x}_{prv} = Dec_{prv}(m_{\theta}(Enc_{prv}(x_{prv})))              \label{equ:final_ds1_samples} \hspace{2mm} where, x_{prv} \in X_{prv} 
    \end{equation}

    \begin{equation}
	\hat{x}_{pub} = Dec_{pub}(m_{\theta}(Enc_{pub}(x_{pub})))  \label{equ:final_ds2_samples} \hspace{2mm} where, x_{pub} \in X_{pub} 
    \end{equation}
    
    $\hat{x}_{prv}$ and $\hat{x}_{pub}$ are reconstructions of $x_{prv}$ and $x_{pub}$, and the loss is calculated as below
    
    \begin{equation}
        \text{Loss} = \frac{1}{n_{prv}} \sum_{d=1}^{D_{prv}} (x^d_{prv} - \hat{x}^d_{prv})^{2} + \frac{1}{n_{pub}} \sum_{d=1}^{D_{pub}}
    (x^d_{pub} - \hat{x}^d_{pub})^{2}
        \label{equ:reconstruction_loss_details}
    \end{equation}
    
    In the inference phase, the model transforms private dataset samples into a format resembling the public dataset. This is done by passing private data through its own encoder (Encoder$_{prv}$) and the shared layers ($m_{\theta}$), followed by reconstruction through the public dataset decoder (Decoder$_{pub}$), producing "crossover" samples.
    \begin{equation}
	\hat{x}_{co} = Dec_{pub}(m_{\theta}(Enc_{prv}(x_{prv})))  \label{equ:final_crossover_ds1} \hspace{2mm} where, x_{prv} \in X_{prv} 
    \end{equation}
    
    This approach enables outlier detection by aligning the representations of private and public data, facilitating the transformation of private dataset outliers into a public dataset format. To evaluate the success of Dataset Transformation, we use the  "DS Diff" Metric \cite{Herurkar-ijcnn23}. Using this metric, we compare the private dataset to both the public and transformed crossover datasets. 
    A lower "DS Diff" value between the private and crossover datasets (compared to the "DS Diff" between the private and public datasets) indicates that the crossover dataset is more similar to the private dataset. 
    \begin{equation}
        \text{DS\_Diff}
    (X_{prv}, X_{pub}) > \text{DS\_Diff}(X_{prv}, \hat{X}_{co})
	\label{equ:hypo2}
    \end{equation}
    This demonstrates that the transformation successfully aligns the crossover dataset with the private dataset while preserving public dataset characteristics.
\begin{figure*}
\centering
\includegraphics[width=\linewidth]{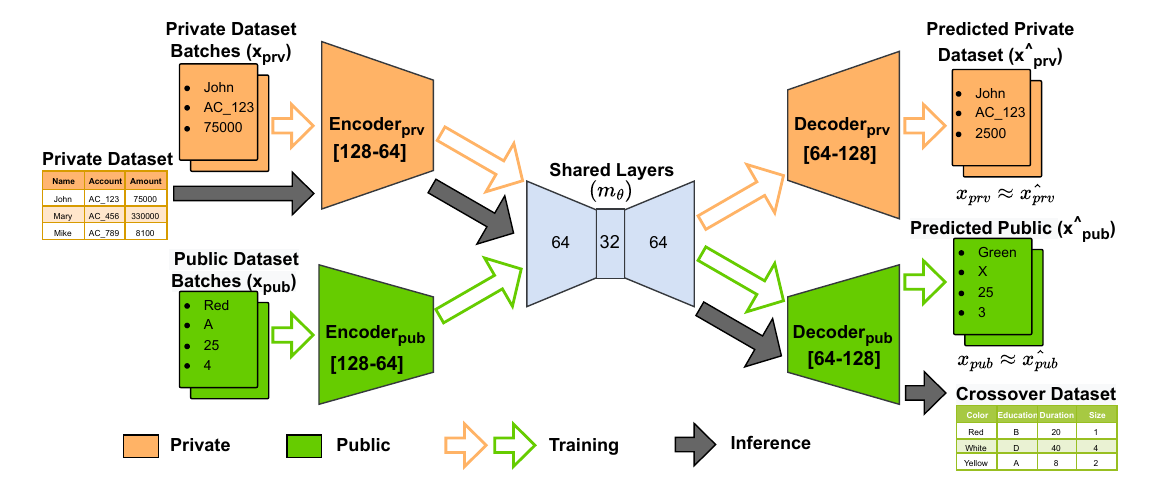}
\caption{Dataset Transformation Model: The model comprises dedicated encoder-decoder pairs for private and public datasets to learn dataset-specific patterns, and shared layers to capture common patterns across both datasets. During inference, private dataset samples are encoded, processed through shared layers, and reconstructed using the public dataset decoder to produce "crossover" samples. This transformation aligns private data with the public dataset format that enables the direct application of optimized public dataset models on crossover samples. \vspace{2mm}}
\label{fig:ds_transformation_model}
\end{figure*}

    \item \textbf{Soft labels prediction and fusion:} 
    After transforming the private dataset into a format similar to the public dataset using the previous components, the next step involves predicting labels for the transformed data. By leveraging prior research or conducting an exhaustive search, we have already identified the best-performing models for each public dataset. 
    These pre-existing, optimized models are then applied to the transformed private dataset (or "crossover" samples) to generate soft labels. In cases where multiple public models are used, we employ fusion techniques such as majority voting to combine their predictions into a final set of soft labels. This ensures robustness and improves the accuracy of the label predictions. Finally, these soft labels are mapped back to the original private dataset, classifying samples as either normal instances or outliers.
\end{itemize}

\subsection{Variations of our Soft Labeling Model:}  \label{sec:approaches}
We propose two variations of the soft labeling approach to handle unlabeled private datasets for outlier detection. Both approaches leverage the "Dataset Similarity Measure" and "Dataset Transformation" techniques, but differ in their strategy for selecting and utilizing public datasets and models. These variations are described below.
\begin{itemize}
    \item \textbf{Method1 (Top1-DS):} 
    In this approach, we aim to identify the most similar public dataset to the private dataset and use it for label generation. First, we compute the similarity between the private dataset and all available public datasets using the "Dataset Similarity Measure". The public dataset with the highest similarity is selected for further steps. We then apply the "Dataset Transformation" to align the private dataset with the selected public dataset's format.
    If the transformation is successful (\autoref{equ:hypo2}), we choose the best $m$ optimized models 
    that perform best on the public dataset and apply them to the transformed (crossover) private dataset.
    The predictions from these models are combined using majority voting to generate soft labels, which are then mapped back to the original private dataset, classifying the samples as either normal or outliers.
    If the transformation fails, the next most similar public dataset is selected, and the process is repeated until a successful transformation is achieved.
    This method ensures efficient soft labeling by leveraging the closest matching public dataset and its best-performing models.
    
    \item \textbf{Method2 (TopN-DS):} 
    In the second approach, we focus on leveraging multiple similar public datasets rather than relying on just one. After computing the similarity between the private dataset and all available public datasets using the "Dataset Similarity Measure", we begin transforming the private dataset into the format of the most similar public datasets. This transformation process continues until successful transformations are achieved for the top $n$ public datasets. For each public dataset where transformation is successful, the best-performing model 
    is selected. Predictions from the transformed (crossover) private datasets are then generated by these models, and a majority voting strategy is used to fuse these predictions into a unified set of soft labels.
    These labels are then mapped back to the original private dataset.
    By using multiple similar public datasets and their best model each, this method can enhance robustness and accuracy in generating soft labels for the private dataset.
\end{itemize}

\section{Experimental Setup}
This section includes descriptions of the datasets and data pre-processing procedures, alongside the baseline methods used for comparison, and different evaluation metrics.
All experiments involving "Dataset Transformation" were implemented using PyTorch v$2$ \cite{pytorch}. 
Each transformation model with Leaky Relu activation function was trained for 1000 epochs using the Adam optimizer \cite{Adam} with parameters $\beta_{1}=0.9$, $\beta_{2}=0.999$, and learning rate=0.001.
The architecture for the "Dataset Transformation" module is structured as Encoder$_{prv}$:[128,64], Encoder$_{pub}$:[128,64], Shared Layers:[64, 32, 64], Decoder$_{prv}$:[64,128], Decoder$_{pub}$:[64,128].
This model setup enables learning a common latent space between structurally different datasets, facilitating effective transformation and soft label generation.

\subsection{Datasets:}   \label{sec:datasets} 
For our study, we utilize a comprehensive selection of datasets, primarily sourced from ADBench \cite{han2022adbench}, one of the most extensive benchmarks for tabular outlier detection. 
ADBench evaluates the performance of many outlier detection algorithms across 57 datasets to assess their effectiveness. 
Of these, 47 commonly used real-world tabular datasets were selected, representing diverse fields such as healthcare, chemistry, astronautics, linguistics, biology, sociology, botany, and finance.
The remaining 10 datasets were excluded as they were either synthetically generated or primarily consisted of images. 
To enhance our analysis, we also include three additional financial datasets: "Credit Default \cite{CD}", which includes data on credit card bill statements, default payments, and demographic details of clients in Taiwan; "Adult Data \cite{AD}", a dataset with personal income records used to predict whether an individual’s income exceeds \$50,000 annually; and "Statlog \cite{UCI}", containing German credit data. 
In total, 50 tabular datasets are employed in our experiments, focusing on outlier detection. 

\subsection{Baseline Methods:}   \label{sec:baseline}
To evaluate the effectiveness of our approach, we conducted a comparative analysis against three baseline methods:
\begin{itemize}
    \item \textbf{Avg OD:} This baseline leverages 11 standard unsupervised OD algorithms from \cite{han2022adbench}, including HBOS, Isolation Forest, CBLOF, ABOD, LOF, OCSVM, MCD, PCA, KNN, Feature Bagging (FB), and Autoencoder (AE). Following the standard procedure outlined in \cite{goldstein}, each OD algorithm was run with multiple hyperparameter configurations. The average performance across these configurations was calculated to assess soft labeling accuracy on the datasets.
    \item \textbf{Default OD:} In this baseline, the same 11 OD algorithms were executed using their default hyperparameter settings. The resulting soft label prediction accuracy provides a reference for performance without hyperparameter tuning.
    \item \textbf{Best OD:} This baseline explores the optimal performance achievable by each OD algorithm. Using grid search, we optimized hyperparameters for each algorithm and selected the best configuration to generate soft labels. While this method sets an upper bound on soft labeling accuracy, it assumes the availability of labeled datasets, which is unrealistic in most practical scenarios. 
\end{itemize}


\subsection{Evaluation Metrics:} 
To assess the effectiveness of our approach, we utilized four key metrics to evaluate labeling performance. 
Since outlier detection datasets are often imbalanced with a majority of inliers, we employed "Balanced Accuracy" across classes to ensure balanced performance measurement, preventing bias towards the dominant class. 
Additionally, we used the "F1-Score", Precision-Recall Area Under the Curve ("PR-AUC"), and Receiver Operating Characteristic Area Under the Curve ("ROC-AUC"), which are widely recognized metrics for evaluating the detection (marking) of outlier samples. 
These metrics provided a comprehensive evaluation of the labeling performance, particularly in identifying outliers.


\section{Experiments and Results}
This section provides a detailed overview of the experiments conducted, offering insights into the performance of our approach through various evaluation metrics and a thorough analysis of the results for each experiment.
The code and models used in this study will be publicly released upon the publication of this paper to facilitate reproducibility and further research.

\subsection{Soft Labeling Performance:} \label{subsec: soft_label_per}
In this experiment, we evaluate the accuracy of the soft labels generated by our approach against baseline models. 
For each of the 50 datasets, one dataset is treated as unlabeled private dataset while the remaining 49 serve as labeled public datasets in a leave one out fashion. 
Using the "Dataset Similarity Measure", we calculate the similarity between the private dataset and the public ones.
For Method1, we transform the private samples to the most similar public dataset using "Dataset Transformation" and apply the best $m$ public models to generate soft label predictions, followed by majority voting to finalize the label prediction. 
In Method2, we perform "Dataset Transformation" to the top $n$ most similar public datasets, applying the best model for each, and again use majority voting to obtain the soft labels.
Baseline models as described above are also used to generate soft labels for each private dataset. 
The predicted soft labels are then compared against the ground truth labels, and the results across all 50 datasets are averaged.

\begin{table}[tbh]
\centering
\caption{Performance metrics and computation time for Method1 and Method2 across 50 datasets, including Balanced Accuracy, F1-score, PR-AUC, and ROC-AUC. Results highlight superior labeling accuracy compared to baseline models, with lower computational time. \vspace{2mm}}
\begin{tabular}{@{}lccccr@{}}
\toprule
\multicolumn{1}{c}{\textbf{Approach}} & \textbf{\begin{tabular}[c]{@{}c@{}}Balanced \\ Accuracy\end{tabular}} & \textbf{\begin{tabular}[c]{@{}c@{}}F1\\ Score\end{tabular}} & \textbf{\begin{tabular}[c]{@{}c@{}}PR\\ AUC\end{tabular}} & \textbf{\begin{tabular}[c]{@{}c@{}}ROC\\ AUC\end{tabular}} & \multicolumn{1}{c}{\textbf{\begin{tabular}[c]{@{}c@{}}Time \\ (mins)\end{tabular}}} \\ \midrule
Best OD (supervised) & 0.722 & 0.487 & 0.470 & 0.802 & 274.38 \\
Avg OD & 0.578 & 0.173 & 0.330 & 0.691 & 282.64 \\
Default OD & 0.577 & 0.170 & 0.331 & 0.690 & 27.72 \\
\textbf{Method1 (Ours)} & \textbf{0.639} & \textbf{0.228} & \textbf{0.477} & \textbf{0.689} & \textbf{9.94} \\
\textbf{Method2 (Ours)} & \textbf{0.625} & \textbf{0.235} & \textbf{0.486} & \textbf{0.683} & \textbf{27.28} \\ \bottomrule
\end{tabular}
\label{tab:method1_method2_per}
\end{table}

\begin{figure*}[t!]
\centering
\includegraphics[width=\linewidth]{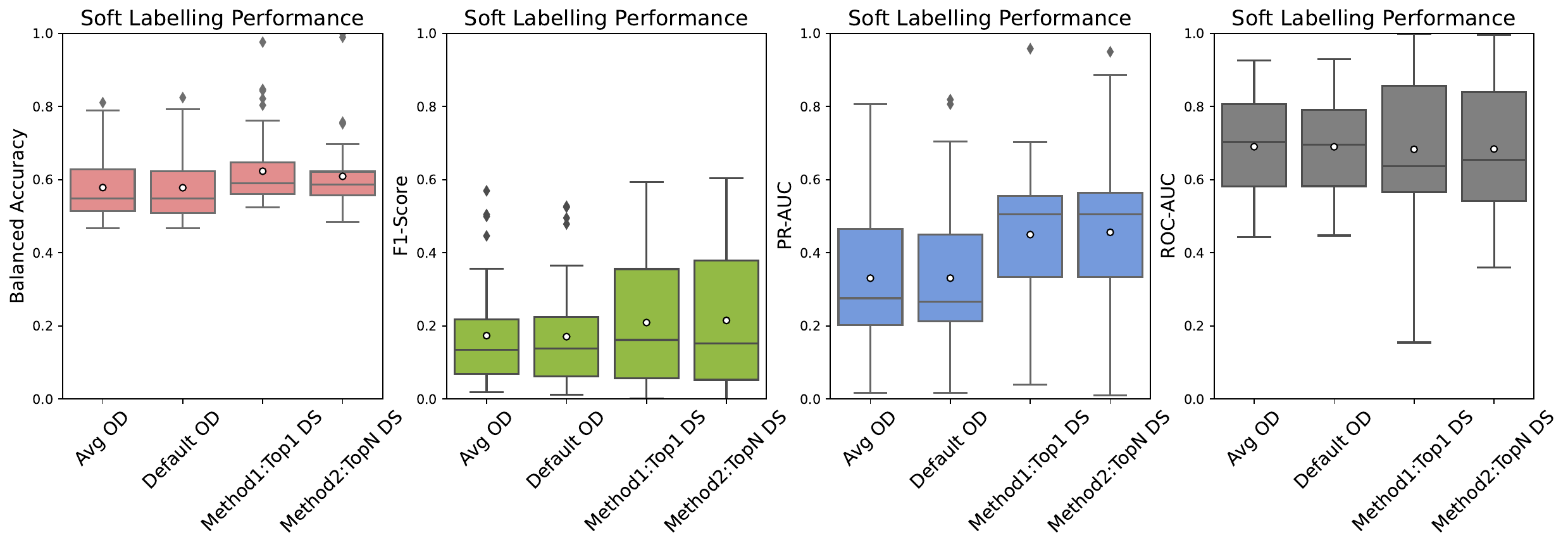}
\caption{Distribution of performance metrics (Balanced Accuracy, F1-score, PR-AUC, and ROC-AUC) across 50 datasets for both methods compared to baseline models. The white dot in each plot represents the mean value. The results demonstrate consistent superiority of the proposed methods in most metrics, with Method 1 and Method 2 showing comparable performance. \vspace{2mm}}
\label{fig:res_distr_plot}
\end{figure*}

\autoref{tab:method1_method2_per} presents the detailed performance results for Method1 and Method2 across all evaluation metrics. 
Both of our methods outperform Avg OD and Default OD baseline models and only fall short of the unrealistic (because of lacking labels) Best OD baseline in terms of Balanced Accuracy, F1-score, and PR-AUC.
This suggests that our approach is able to generate more accurate soft labels.
However, while our methods excel in most metrics, the ROC-AUC values are lower compared to the baseline models. 
This could be due to the inherent noise introduced during dataset transformation, or differences in the model’s sensitivity when dealing with borderline cases of outliers and inliers.
Despite this, both Method1 and Method2 perform similarly, showing that the use of either a single most similar dataset or multiple similar datasets yields comparable results in soft label creation. 
Additionally, we analyzed the overall distribution of performance across all 50 datasets and plotted the results in \autoref{fig:res_distr_plot}. 
The distribution plots clearly support the conclusion that our approach outperforms baseline models, reinforcing the findings from \autoref{tab:method1_method2_per}.  
These results demonstrate that our method not only enhances the labeling accuracy but also offers a reliable solution for creating soft labels in the absence of true labels, especially in complex outlier detection scenarios.

\subsection{Compute Time:}
The computational time required for each method across all datasets is summarized in \autoref{tab:method1_method2_per}.  
On average, both Method1 and Method2 are significantly more efficient compared to the baseline approaches. 
Notably, Method1 demonstrates a markedly lower computational time, which is due to the fact that it typically requires only a single dataset transformation. 
In contrast, Method2 involves multiple transformations, which naturally increases its processing time. 
This difference is particularly evident when handling larger datasets.

\begin{figure}
\centering
\includegraphics[width=\linewidth]{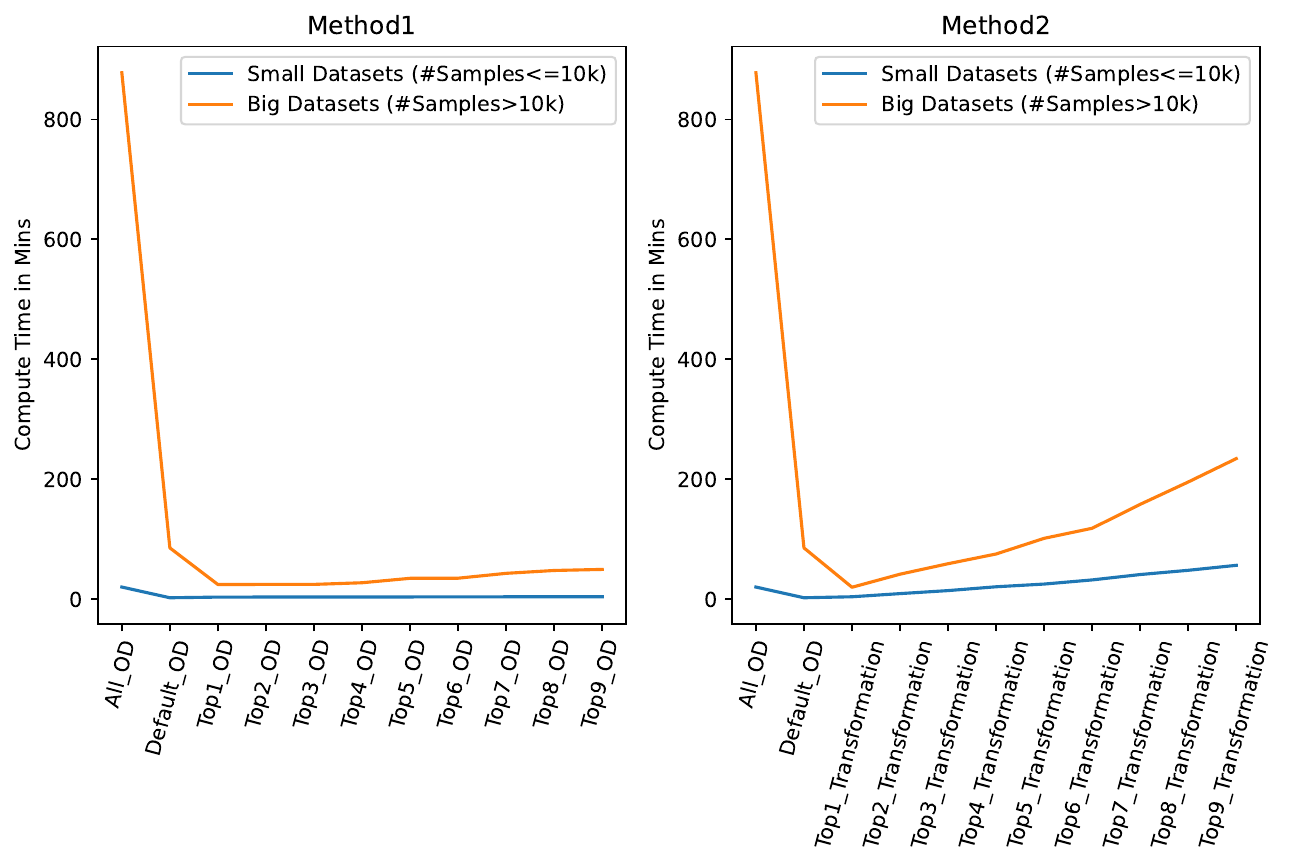}
\caption{
Impact of best $m$ model and top $n$ transformation selection on computation time for both methods.
While (single threaded) computation time for Method1 remains largely unaffected by increasing $m$, Method2 shows significant runtime increases, especially for datasets with over 10,000 samples, reflecting the trade-off between accuracy and efficiency. \vspace{4mm}}
\label{fig:top_n_time}
\end{figure}
Additionally, we analyzed the effect of selecting best $m$ models and top $n$ transformations on computation time for both methods. 
As illustrated in \autoref{fig:top_n_time}, increasing the value of $m$ does not substantially impact the runtime of Method 1. 
However, for Method 2, the computation time increases with $n$, particularly for datasets with more than 10,000 samples. 
This shows that while Method 2 offers the potential for improved accuracy by leveraging multiple dataset transformations, it comes with the trade-off of higher computational costs when dealing with larger datasets. 

\subsection{Discussion:}
\begin{figure}[htb]
\centering
\includegraphics[width=\linewidth]{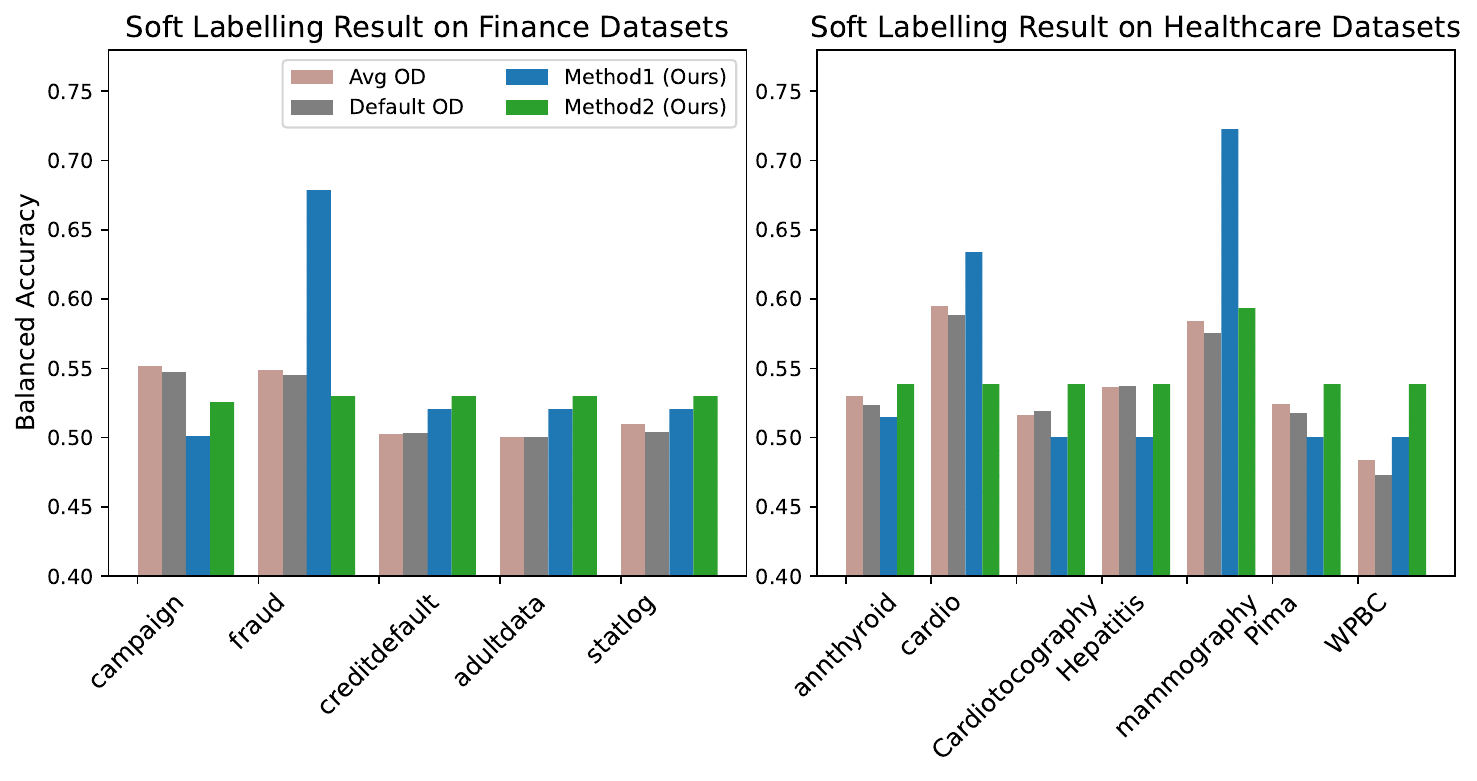}
\caption{Soft label accuracy results for individual datasets in the healthcare and finance domains. The x-axis shows the held out unlabeled (private) datasets, and the y-axis shows performance scores. Both Method1 and Method2 consistently outperform baseline methods across individual datasets, demonstrating their robustness in varied domains. \vspace{4mm}}
\label{fig:ds_wise_res}
\end{figure}

\noindent \textbf{Individual Dataset Result:}
To further understand the performance of soft label generation, we conducted an ablation study by analyzing results at the individual dataset level. 
Specifically, we grouped datasets into healthcare and finance domains and examined the soft label accuracy for each dataset within these groups. 
\autoref{fig:ds_wise_res} presents the results, with the x-axis representing the names of the held out unlabeled (private) datasets and the y-axis showing the performance scores. 
These results demonstrate that both Method 1 and Method 2 outperform baseline methods consistently, even when evaluated on individual datasets.

\begin{figure}[htb]
\centering
\includegraphics[width=\linewidth]{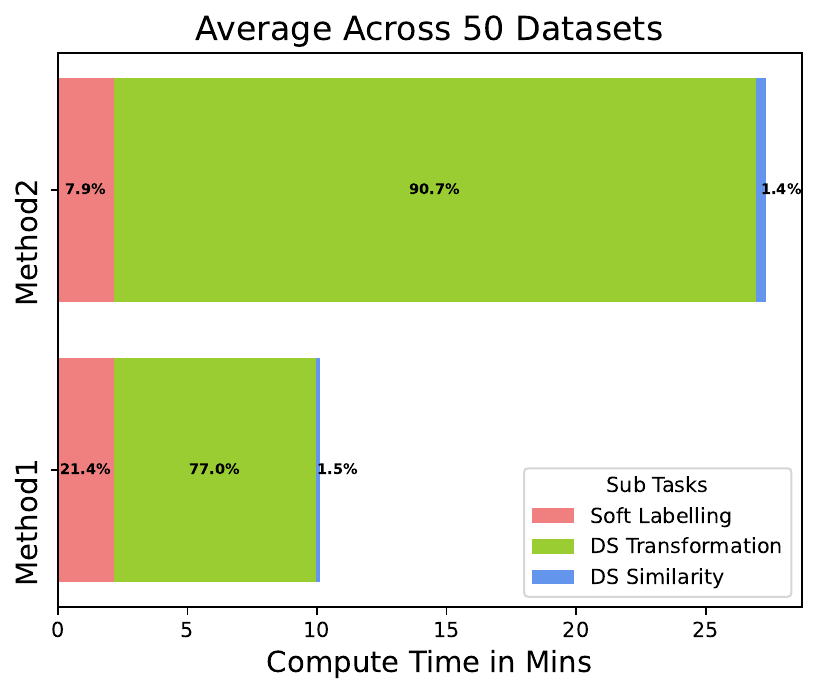}
\caption{Breakdown of (single threaded) computational time for different components of the proposed methods. The dataset transformation step accounts for the majority of the total computation time, particularly in Method2, where multiple transformations are required. The figure highlights the potential for optimization through parallelization to reduce compute time. \vspace{4mm}}
\label{fig:opt_time}
\end{figure}

\noindent \textbf{Possible Compute Time Optimization:}
Additionally, we analyzed the computational time required for different components of our approach, as shown in \autoref{fig:opt_time}. 
It is evident from the figure that the bulk of the total computation time is consumed by the "Dataset Transformation" step. 
This observation highlights an opportunity for optimization, particularly in Method2, where multiple dataset transformations are required. 
By parallelizing the "Dataset Transformation" process, it would be possible to significantly reduce the compute time for Method2, potentially bringing it down to the levels of Method1 in the best-case scenario.

\begin{figure}[t!]
\centering
\includegraphics[width=\linewidth]{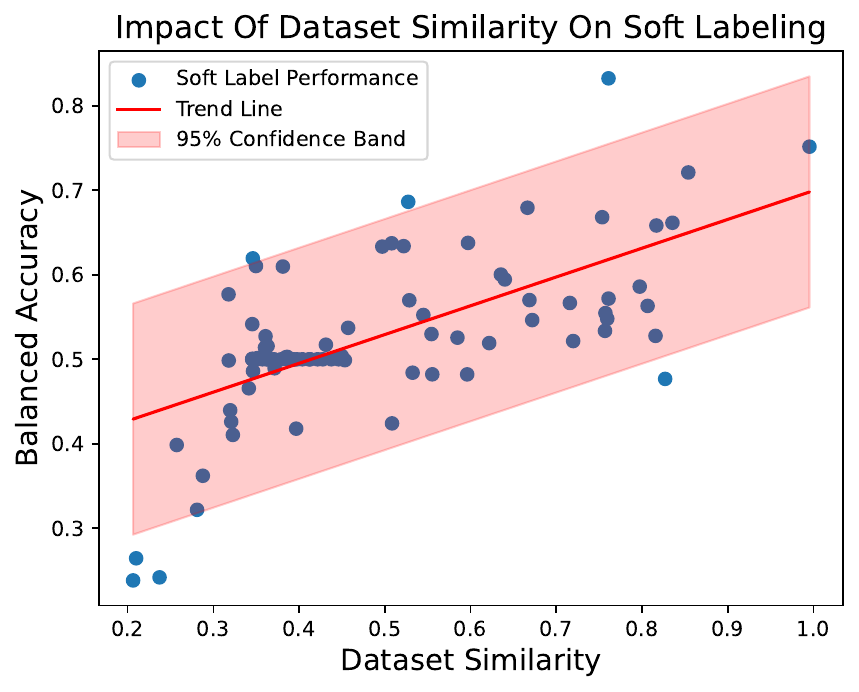}
\caption{Relationship between dataset similarity and soft labeling performance. The x-axis represents the similarity scores between unlabeled and labeled datasets, while the y-axis shows the balanced accuracy of soft label predictions. The upward trend demonstrates that higher dataset similarity leads to improved soft labeling performance, highlighting the importance of selecting similar datasets for transformation and prediction. \vspace{4mm}}
\label{fig:sim_vs_softlabeling}
\end{figure}

\noindent \textbf{Impact of Dataset Similarity on Soft Labeling:}
To better understand how dataset similarity affects soft labeling—specifically, the selection of labeled datasets based on their similarity to an unlabeled dataset—we conducted an in-depth analysis. 
Using the "Dataset Similarity Measure", we identified the most similar labeled dataset for each of these from the available dataset list in \autoref{sec:datasets}.
We then applied Method1 (\autoref{sec:approaches}) to transform each unlabeled dataset into the identified similar datasets and predicted soft labels for each transformation. 
The relationship between dataset similarity and soft labeling performance is visualized in \autoref{fig:sim_vs_softlabeling}. 
The x-axis represents the similarity score between datasets, while the y-axis shows the balanced accuracy of the soft label predictions.
The results in \autoref{fig:sim_vs_softlabeling} reveal an upward trend: as dataset similarity increases, the balanced accuracy of soft label predictions also improves. 
This indicates that selecting highly similar datasets using the "Dataset Similarity Measure" is critical for effective transformation and accurate soft label generation. 
Thus, dataset similarity plays a pivotal role in enhancing the quality of soft labeling.


\section{Conclusion}
In this paper, we introduced a novel approach for generating soft labels in outlier detection tasks by transforming private datasets into formats resembling public datasets. 
Our method includes two key components: a "Dataset Similarity Measure" and a "Dataset Transformation" model. 
We developed two variations of the soft labeling approach—Method1, which uses the most similar public dataset, and Method2, which leverages multiple similar public datasets with the best available models for label prediction. 
Both methods were evaluated across 50 tabular datasets, including those from the finance and healthcare domains, and consistently outperformed baseline models in terms of balanced accuracy, F1 score, and PR-AUC.
Our experiments showed that Method1, requiring fewer dataset transformations, was computationally more efficient than Method2, which involved multiple transformations. 
Nonetheless, the potential for parallelization in Method2 could bring its computation time on par with Method1. 
Additionally, an ablation study on individual datasets demonstrated that both methods frequently performed better than baselines, reinforcing the versatility of our approach across different domains.

In conclusion, our approach offers a robust, scalable, and novel solution for generating accurate soft labels for unlabeled private datasets using existing public models, providing a valuable tool for outlier detection and improving the overall performance in diverse real-world scenarios. 
Further optimization in computational time, especially for Method2, presents opportunities for future work in improving efficiency.
Also, incorporating better fusion strategies can become beneficial in the long run.








\newpage

\bibliography{mybibfile}

\newpage

\onecolumn
\thispagestyle{empty} 
\vspace*{\fill}
\begin{center}
    {\Huge \textbf{Appendix}}
\end{center}
\vspace*{\fill}
\twocolumn

\begin{table*}[htbp]
\begin{tabular}{@{}lrrrrlrrrrrr@{}}
\toprule
\multicolumn{1}{c}{\multirow{2}{*}{\textbf{Dataset}}} & \multicolumn{1}{c}{\multirow{2}{*}{\textbf{Samples}}} & \multicolumn{1}{c}{\multirow{2}{*}{\textbf{Features}}} & \multicolumn{1}{c}{\multirow{2}{*}{\textbf{Outliers}}} & \multicolumn{1}{c}{\multirow{2}{*}{\textbf{\begin{tabular}[c]{@{}c@{}}Outlier\\ \%\end{tabular}}}} & \multicolumn{1}{c}{\multirow{2}{*}{\textbf{\begin{tabular}[c]{@{}c@{}}Best OD\\ Algo\end{tabular}}}} & \multicolumn{3}{c}{\textbf{PRAUC}} & \multicolumn{3}{c}{\textbf{ROCAUC}} \\ \cmidrule(l){7-12} 
\multicolumn{1}{c}{} & \multicolumn{1}{c}{} & \multicolumn{1}{c}{} & \multicolumn{1}{c}{} & \multicolumn{1}{c}{} & \multicolumn{1}{c}{} & \textbf{Best OD} & \textbf{Default OD} & \textbf{Avg OD} & \textbf{Best OD} & \textbf{Default OD} & \textbf{Avg OD} \\ \midrule
1\_ALOI & 49534 & 27 & 1508 & 3.04 & ABOD & 0.1824 & 0.0660 & 0.0581 & 0.7608 & 0.6036 & 0.5920 \\
2\_annthyroid & 7200 & 6 & 534 & 7.41 & MCD & 0.3868 & 0.2620 & 0.2636 & 0.8366 & 0.7373 & 0.7214 \\
3\_backdoor & 95329 & 196 & 2329 & 2.44 & PCA & 0.5326 & 0.2449 & 0.2628 & 0.8781 & 0.7424 & 0.7663 \\
4\_breastw & 683 & 9 & 239 & 34.99 & I\_Forest & 0.9865 & 0.8190 & 0.8073 & 0.9931 & 0.8495 & 0.8231 \\
5\_campaign & 41188 & 62 & 4640 & 11.26 & HBOS & 0.4061 & 0.2442 & 0.2559 & 0.7993 & 0.6964 & 0.6965 \\
6\_cardio & 1831 & 21 & 176 & 9.61 & I\_Forest & 0.6434 & 0.3964 & 0.4094 & 0.9517 & 0.7409 & 0.7729 \\
7\_Cardiotocography & 2114 & 21 & 466 & 22.04 & OCSVM & 0.6058 & 0.3823 & 0.3843 & 0.5061 & 0.6066 & 0.6062 \\
8\_celeba & 202599 & 39 & 4547 & 2.24 & MCD & 0.1177 & 0.0591 & 0.0626 & 0.8364 & 0.6471 & 0.6600 \\
9\_census & 299285 & 500 & 18568 & 6.21 & ABOD & 0.5074 & 0.1406 & 0.1322 & 0.6576 & 0.6084 & 0.6275 \\
10\_cover & 286048 & 10 & 2747 & 0.96 & OCSVM & 0.5048 & 0.1017 & 0.1199 & 0.5000 & 0.7648 & 0.7763 \\
11\_donors & 619326 & 10 & 36710 & 5.92 & CBLOF & 0.2365 & 0.1495 & 0.1445 & 0.8904 & 0.7350 & 0.7163 \\
12\_fault & 1941 & 27 & 673 & 34.67 & ABOD & 0.5279 & 0.4169 & 0.4076 & 0.7241 & 0.5835 & 0.5795 \\
13\_fraud & 284807 & 29 & 492 & 0.17 & ABOD & 0.5011 & 0.2146 & 0.2147 & 0.5941 & 0.8490 & 0.8126 \\
14\_glass & 214 & 7 & 9 & 4.21 & ABOD & 0.2008 & 0.1086 & 0.1001 & 0.8656 & 0.6952 & 0.6940 \\
15\_Hepatitis & 80 & 19 & 13 & 16.25 & OCSVM & 0.5833 & 0.2681 & 0.2790 & 0.5000 & 0.5788 & 0.5819 \\
16\_http & 567498 & 3 & 2211 & 0.38 & I\_Forest & 0.6790 & 0.2369 & 0.2535 & 0.9998 & 0.6795 & 0.7324 \\
17\_InternetAds & 1966 & 1555 & 368 & 18.71 & MCD & 0.5947 & 0.4414 & 0.4333 & 0.8050 & 0.6879 & 0.6695 \\
18\_Ionosphere & 351 & 32 & 126 & 35.89 & MCD & 0.9470 & 0.8065 & 0.7847 & 0.9539 & 0.8467 & 0.8220 \\
19\_landsat & 6435 & 36 & 1333 & 20.71 & CBLOF & 0.3071 & 0.2224 & 0.2168 & 0.6292 & 0.5156 & 0.5089 \\
20\_letter & 1600 & 32 & 100 & 6.25 & OCSVM & 0.4443 & 0.2307 & 0.2090 & 0.8693 & 0.7371 & 0.7254 \\
21\_Lymphography & 148 & 18 & 6 & 4.05 & AE & 1.0000 & 0.5571 & 0.5471 & 1.0000 & 0.8884 & 0.9114 \\
22\_magic.gamma & 19020 & 10 & 6688 & 35.16 & ABOD & 0.7384 & 0.6167 & 0.6118 & 0.8138 & 0.7044 & 0.6986 \\
23\_mammography & 11183 & 6 & 260 & 2.32 & I\_Forest & 0.2192 & 0.1381 & 0.1364 & 0.8682 & 0.8048 & 0.8125 \\
24\_mnist & 7603 & 100 & 700 & 9.20 & OCSVM & 0.5460 & 0.3607 & 0.3469 & 0.5000 & 0.7710 & 0.7492 \\
25\_musk & 3062 & 166 & 97 & 3.16 & AE & 1.0000 & 0.5253 & 0.5118 & 1.0000 & 0.6218 & 0.7076 \\
26\_optdigits & 5216 & 64 & 150 & 2.87 & OCSVM & 0.4033 & 0.1192 & 0.0778 & 0.3893 & 0.5739 & 0.5124 \\
27\_PageBlocks & 5393 & 10 & 510 & 9.45 & AE & 0.5790 & 0.4032 & 0.4132 & 0.9202 & 0.7502 & 0.7614 \\
28\_pendigits & 6870 & 16 & 156 & 2.27 & PCA & 0.2117 & 0.1134 & 0.1111 & 0.9459 & 0.7864 & 0.7879 \\
29\_Pima & 768 & 8 & 268 & 34.89 & OCSVM & 0.6753 & 0.4670 & 0.4654 & 0.5000 & 0.5853 & 0.5805 \\
30\_satellite & 6435 & 36 & 2036 & 31.63 & MCD & 0.7687 & 0.6072 & 0.5919 & 0.8054 & 0.6576 & 0.6399 \\
31\_satimage-2 & 5803 & 36 & 71 & 1.22 & AE & 0.9535 & 0.4480 & 0.4882 & 0.9957 & 0.7516 & 0.8083 \\
32\_shuttle & 49097 & 9 & 3511 & 7.15 & I\_Forest & 0.9804 & 0.5560 & 0.5739 & 0.9975 & 0.8073 & 0.8052 \\
33\_skin & 245057 & 3 & 50859 & 20.75 & MCD & 0.4732 & 0.2619 & 0.2528 & 0.8830 & 0.6191 & 0.5992 \\
34\_smtp & 95156 & 3 & 30 & 0.03 & CBLOF & 0.6680 & 0.2110 & 0.1768 & 0.9746 & 0.8856 & 0.8685 \\
35\_SpamBase & 4207 & 57 & 1679 & 39.91 & OCSVM & 0.6129 & 0.4546 & 0.4684 & 0.7069 & 0.5638 & 0.5819 \\
36\_speech & 3686 & 400 & 61 & 1.65 & CBLOF & 0.0344 & 0.0175 & 0.0164 & 0.6471 & 0.4887 & 0.4784 \\
37\_Stamps & 340 & 9 & 31 & 9.11 & HBOS & 0.4698 & 0.3178 & 0.3203 & 0.9486 & 0.8108 & 0.8169 \\
38\_thyroid & 3772 & 6 & 93 & 2.46 & MCD & 0.6948 & 0.3911 & 0.4164 & 0.9848 & 0.8767 & 0.8798 \\
39\_vertebral & 240 & 6 & 30 & 12.50 & OCSVM & 0.5625 & 0.1665 & 0.1811 & 0.5000 & 0.4979 & 0.5161 \\
40\_vowels & 1456 & 12 & 50 & 3.43 & KNN & 0.6974 & 0.2632 & 0.2727 & 0.9615 & 0.7698 & 0.7624 \\
41\_Waveform & 3443 & 21 & 100 & 2.90 & KNN & 0.3011 & 0.0928 & 0.1080 & 0.8425 & 0.6751 & 0.6677 \\
42\_WBC & 223 & 9 & 10 & 4.48 & MCD & 0.9028 & 0.4893 & 0.5399 & 0.9948 & 0.8800 & 0.8893 \\
43\_WDBC & 367 & 30 & 10 & 2.72 & CBLOF & 1.0000 & 0.7037 & 0.7251 & 1.0000 & 0.9290 & 0.9262 \\
44\_Wilt & 4819 & 5 & 257 & 5.33 & OCSVM & 0.5266 & 0.1188 & 0.1220 & 0.5000 & 0.5020 & 0.5211 \\
45\_wine & 129 & 13 & 10 & 7.75 & KNN & 1.0000 & 0.5862 & 0.4878 & 1.0000 & 0.8827 & 0.8117 \\
46\_WPBC & 198 & 33 & 47 & 23.73 & CBLOF & 0.3680 & 0.2130 & 0.2242 & 0.6778 & 0.4895 & 0.5043 \\
47\_yeast & 1484 & 8 & 507 & 34.16 & CBLOF & 0.3836 & 0.3292 & 0.3226 & 0.5529 & 0.4473 & 0.4423 \\
48\_creditdefault & 30000 & 146 & 6636 & 22.12 & HBOS & 0.3511 & 0.2571 & 0.2586 & 0.6134 & 0.5548 & 0.5609 \\
49\_adultdata & 48842 & 108 & 11687 & 23.92 & OCSVM & 0.3199 & 0.2430 & 0.2455 & 0.5222 & 0.4715 & 0.4753 \\
50\_statlog & 1000 & 123 & 300 & 30.00 & OCSVM & 0.4163 & 0.3237 & 0.3311 & 0.5902 & 0.5130 & 0.5182 \\ \bottomrule
\end{tabular}
\caption{List of all 50 datasets and their attributes with best OD algorithms.}
\label{tab:all_ds_infor}
\end{table*}

\clearpage

\end{document}